\documentclass[conference]{IEEEtran}
\IEEEoverridecommandlockouts
\usepackage{cite}
\usepackage{amsmath,amssymb,amsfonts}
\usepackage{algorithmic}
\usepackage{graphicx}
\usepackage{textcomp}
\usepackage{xcolor}

\usepackage{multirow}

\def\BibTeX{{\rm B\kern-.05em{\sc i\kern-.025em b}\kern-.08em
    T\kern-.1667em\lower.7ex\hbox{E}\kern-.125emX}}

\begin{document}

\title{ESC: Evolutionary Stitched Camera \\ Calibration in the Wild}

\author{\IEEEauthorblockN{Grzegorz Rypeść}
\IEEEauthorblockA{\textit{Warsaw University of Technology} \\
\textit{Sport Algorithmics and Gaming}\\
Warsaw, Poland \\
grzegorz.rypesc.dokt@pw.edu.pl}
\and
\IEEEauthorblockN{Grzegorz Kurzejamski}
\IEEEauthorblockA{\textit{Sport Algorithmics and Gaming} \\
Warsaw, Poland \\
g.kurzejamski@sagsport.com}
}

\maketitle

\begin{abstract}
This work introduces a novel end-to-end approach for estimating extrinsic parameters of cameras in multi-camera setups on real-life sports fields. We identify the source of significant calibration errors in multi-camera environments and address the limitations of existing calibration methods, particularly the disparity between theoretical models and actual sports field characteristics. We propose the Evolutionary Stitched Camera calibration (ESC) algorithm to bridge this gap. It consists of image segmentation followed by evolutionary optimization of a novel loss function, providing a unified and accurate multi-camera calibration solution with high visual fidelity. The outcome allows the creation of virtual stitched views from multiple video sources, being as important for practical applications as numerical accuracy. We demonstrate the superior performance of our approach compared to state-of-the-art methods across diverse real-life football fields with varying physical characteristics.
\end{abstract}


\section{Introduction}

Nowadays, many sports facilities install cameras around the venue to register matches and training sessions. Various companies \cite{tracab, sagsport, veo} offer services that analyze videos from such installations and deliver analytical data or various video content, be it live streaming or analytical views with 3D overlays. Devices used for video capture are usually static, wide-angle, high-resolution CCTV or industrial cameras. Registering the game with them is the first step in the computer vision analytical pipeline for sports. 

\begin{figure}[ht]
\centerline{\includegraphics[width=0.48\textwidth]{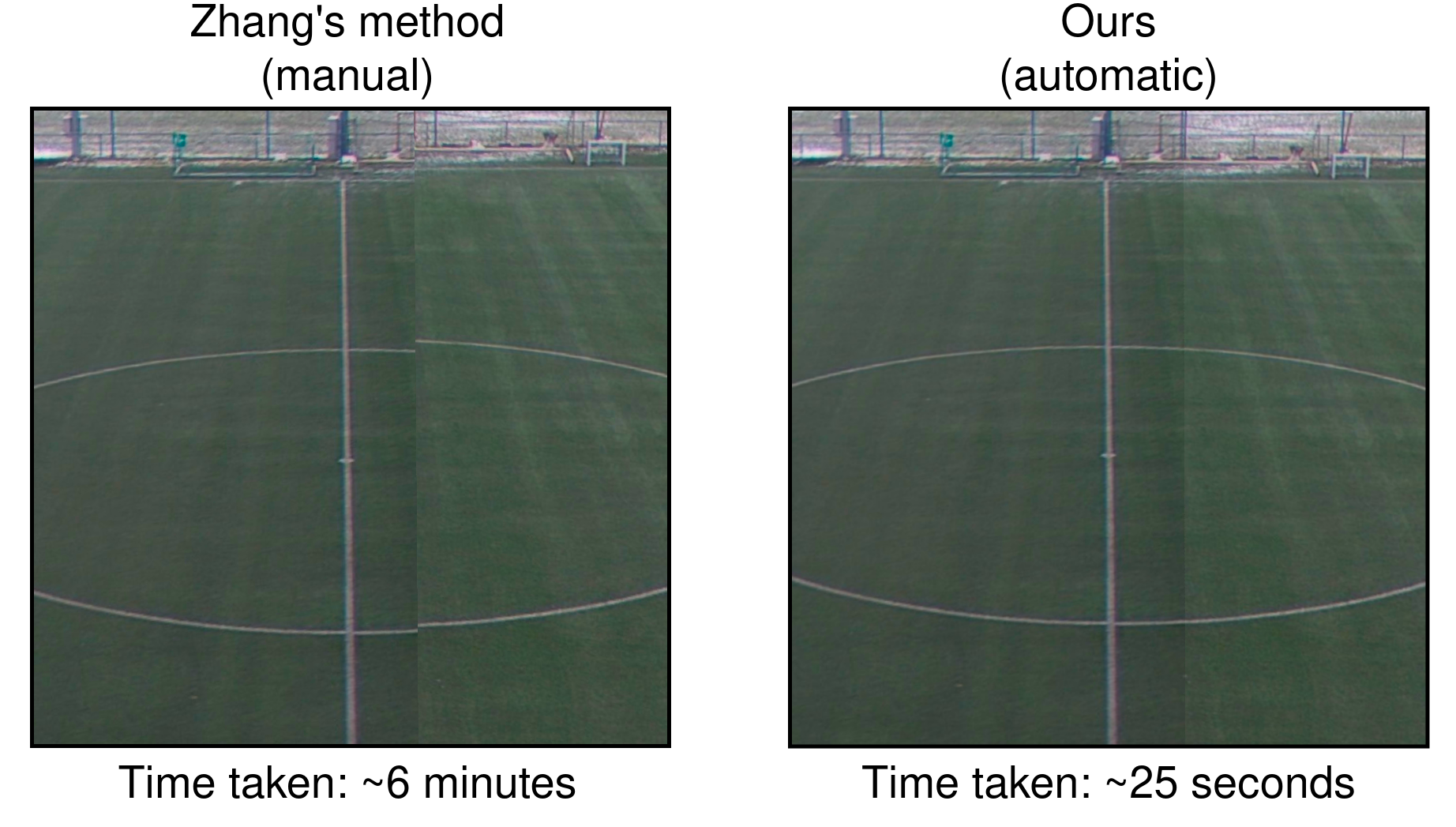}}
\caption{Calibration results for Zhang's and our method. Our approach is faster and presents better stitch and visual fidelity. This is because the latter does not take the stitch into account and assumes constant environmental conditions, e.g., that distortion coefficients are unaffected by the temperature.
}
\label{fig:teaser}
\end{figure}

Analyzing such video streams requires calibration data for each camera. Estimating players' position or detection of game events depends on an ability to map 2D-pixel coordinates in the image plane to 3D coordinates in the world reference frame. Moreover, virtual camera production often uses multiple stitched camera views to obtain higher resolution, as presented in Fig.~\ref{fig:teaser}. Because sports fields usually have low textural cues for dense stitch algorithms, the stitching must be done based on the calibration data. Thus, many calibration methods have been developed to calibrate cameras in regard to the lines or areas of a sports field.

\begin{figure*}[ht]
\centering
\centerline{\includegraphics[width=0.9\textwidth]{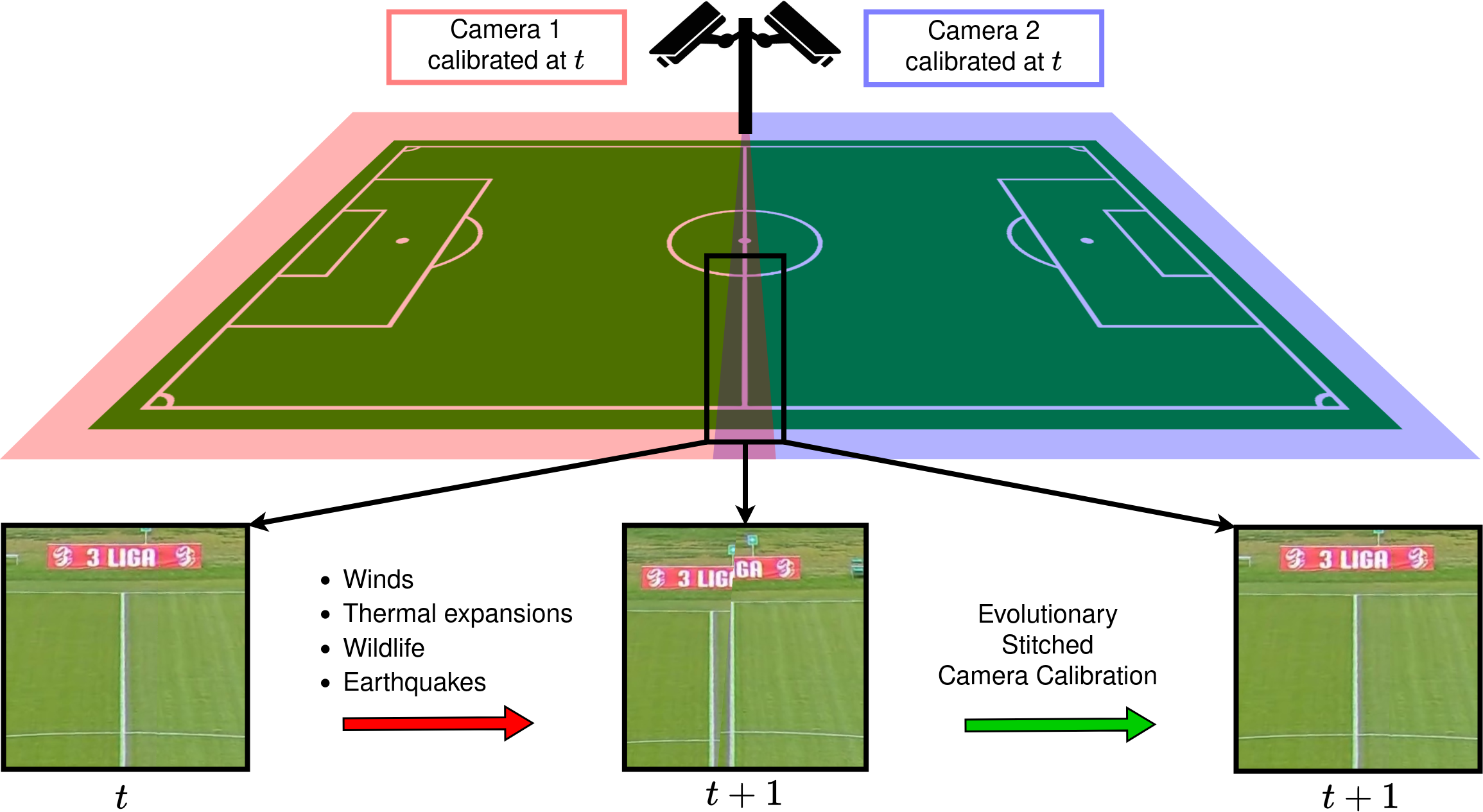}}
\caption{We consider a scenario where two cameras are placed behind the middle line of an outdoor football field and were calibrated at time $t$. Due to adversarial environmental conditions such as winds, thermal expansions, or wildlife, their calibrations are not valid at the $t+1$ moment. This causes errors when stitching videos of these cameras. We utilize our novel ESC method to refine the calibrations and improve the quality of the stitched video. 
}
\label{fig:problem}
\end{figure*}

Unfortunately, cameras installed on high poles and other venues’ infrastructure are not always static. Cameras are affected by adverse environmental conditions as presented in Fig.~\ref{fig:problem}. Strong winds or temperature fluctuations cause small changes to the camera position and viewing angle. These little changes make initial calibration inaccurate over time, which adversely affects the quality of the entire sports analytic pipeline. That imposes real-time demand on the architecture of the algorithm and renders some of the SOTA algorithms unsuited for this task because of computational or time costs.

Additionally, many calibration methods assume the sports field to be planar~\cite{jiang2020optimizing, chen2019sports, rypesc2022sports}. In the case of fields such as football pitches, this is not true in many cases. The center of such areas is designed to be slightly elevated to remove rainwater, as presented in Fig.~\ref{fig:heatmap}. Moreover, slight deviations from the planar surface further adversely impact the calibration tasks.
In virtual production involving a stitched view, additional calibration errors may arise due to the oversight of inherent 3D-related features of the camera installation. When cameras are mounted in the same spot, achieving a perfect physical alignment with a shared projection center point is unattainable. That produces a parallax effect, introducing incoherence in the stitched views in the presence of the non-planar field. That is, in fact, the common occurrence when using algorithms that do not consider the aforementioned issues with non-planar fields and can be observed in Fig.~\ref{fig:stitch}. However, our study demonstrates that a well-designed calibration pipeline can alleviate adverse parallax effects. The result is virtual views where these effects are visually imperceptible, with no significant impact on the numerical accuracy of the 3D positional decoding from the image plane.

To solve the presented issues, we introduce a novel calibration method designed to enable high-quality projections and video stitching with parallax effect compensation mechanisms. It is dubbed ESC (Evolutionary Stitched Camera Calibration). To sum up, our contributions are:
\begin{itemize}
    \item We develop a novel method for camera calibration that combines image segmentation and evolution strategy. It enables high-quality video stitching and projections between image planes and real-world coordinates for real-life scenarios with non-planar sports fields.
    \item We propose a 3D model to estimate football playfields and show that it improves calibration quality over the traditional flat model.
    \item We compare our method to state-of-the-art calibration methods. We show that ESC achieves better quantitative and qualitative results.
\end{itemize}

\section{Related Work}

\subsection{Image stitching}
Image stitching plays a pivotal role in creating cohesive panoramic or high-resolution representations by seamlessly combining overlapping images. Early approaches, such as \cite{szeliski1996video}, apply Levenberg Marquardt (LM) algorithm to increase the quality of stitched panoramic images. To align the foreground and background in panoramic images, the dual-homography warping (DHW) is introduced in \cite{gao2011constructing}. While achieving good results for simple scenes consisting of predominating planes, it performs poorly for more complex scenes. Later, deep neural networks were utilized for the problem, resulting in algorithms such as Learned Invariant Feature Transform (LIFT)~\cite{yi2016lift}. By utilizing convolutional~\cite{lecun1998gradient} and Siamese~\cite{koch2015siamese} neural networks to detect common features in images, the method yields superior results compared to Scale Invariant Feature Transform (SIFT)~\cite{lindeberg2012scale}. \cite{le2020deep} extracts multi-scale image features using image pyramids to predict homographies from coarse to fine. \cite{nie2021unsupervised} reconstructs images' stitched parts, enabling better stitch quality. However, these methods depend on rich and distinctive features of the images that the football field is mostly lacking.


\subsection{Calibrating cameras in sports}
Calibrating cameras in sports is a fundamental step to enable automatic analysis of sports events. Therefore, it is extensively studied in the literature. Early methods, as seen in works like~\cite{wang2006fast, yamada2002tracking, kim2000soccer, farin2003robust}, employ Hough\cite{duda1972use} transforms to extract geometric primitives like curves and ellipses from images. These primitives facilitate the identification of control points or enable a combinatorial search for estimating camera parameters. However, a vast manual effort is required to obtain them, e.g., to select heuristics that recognize certain textures. More advanced methods\cite{puwein2011robust, yang2022sports} utilize SIFT~\cite{lowe2004distinctive} and MSER~\cite{matas2004robust} image features to calibrate cameras, whereas recent methods utilize deep neural networks for automatic feature extraction. \cite{jiang2020optimizing} trains ResNet architecture\cite{he2016deep} to estimate calibration error and utilizes a gradient descent approach to find camera parameters. However, this method tends to get stuck in local minima. Authors of \cite{rypesc2022sports} try to overcome this issue. They utilize a U-Net\cite{ronneberger2015u} architecture to detect playfield lines, which enables the calculation of a fitness function between a warped segmented image and a playfield template. They alleviate the local minima problem by using an evolution strategy to minimize fitness function, thus achieving exact homography matrices. However, the method assumes that the playfield is flat and does not enable good image stitching quality. Generative approaches were also studied. \cite{chen2019sports} utilizes generative adversarial networks~\cite{goodfellow2020generative} (GAN) to detect playfield markings. They initialize the camera pose using a feature-pose database and later improve it based on image distance calculated using the Lucas-Kanade algorithm. However, GAN networks are difficult to train, and the method requires time-consuming preprocessing to obtain the database.

\subsection{Image segmentation}
Image segmentation is the task of assigning a class to each pixel in an image. We tackle it in order to find playfield lines. In this field, deep neural networks achieve superior results compared to older, conventional approaches~\cite{minaee2021image}. One of the first neural architectures designed for image segmentation was~\cite{long2015fully}, consisting only of convolutional layers. \cite{ronneberger2015u} proposed U-Net architecture by adding deconvolution layers and residual connections, which allowed the mix of low-level feature maps with high-level ones. However, these approaches did not take into account global-level information. To solve this issue, \cite{liu2015parsenet} augmented feature maps with their average, thus enabling a global context. The lack of global-level information in convolutional layers was solved by moving from convolutional to visual transformer architectures\cite{dosovitskiy2020image}. Segmentation Transformer (SETR)~\cite{zheng2021rethinking} consists of an encoder and a decoder built from transformer blocks. The encoder transforms the image into a sequence of embeddings while the decoder produces segmentation masks. \cite{strudel2021segmenter} represents segmentation classes as embedding vectors and combines them with image embeddings in the mask transformer decoder. In this work, we utilize a U-Net-based network for segmentation, as it requires less computational resources to infer images in 4K resolution, and global-level information is not necessary in the case of playfield line detection.

\section{Approach}
\subsection{Problem definition}

We define the continual multi-camera extrinsic camera calibration process at time $t \in \{1, 2, ...\}$ as a problem of search in $N \times \mathbb{R}^{6}$ space, where $N$ is the number of cameras involved. The first three dimensions in $n$-th camera ($n \in \{1, 2 ..., N\}$) define its rotation following Rodrigues representation: $\textbf{r}^n_t \in \mathbb{R}^3$. The last three dimensions define a~translation vector $\textbf{l}^n_t \in \mathbb{R}^3$ representing the camera's position in the real-world coordinates. We assume the cameras' intrinsic parameters are constant in time. We estimate them and $\{\textbf{r}^{n}_1, \textbf{l}^{n}_1\}$ at $t=1$, during an initial camera calibration, using a~planar chessboard-like calibration pattern and the Zhang~\cite{zhang2000flexible} method.

We also represent $n$-th camera movement from $t$ to $t+1$ as unknown vector $\{\textbf{r}_{t \rightarrow t+1}^n, \textbf{l}_{t \rightarrow t+1}^n\}$. This movement is undesired and is caused by adversarial environmental conditions. The goal of the continual multi-camera extrinsic camera calibration process is to approximate this movement and find new camera parameters $\{\textbf{r}_{t+1}^n, \textbf{l}_{t+1}^n\}$ so that $\textbf{r}_{t+1}^n = \textbf{r}_{t}^n + \textbf{r}_{t \rightarrow t+1}^n$ and $\textbf{l}_{t+1}^n = \textbf{l}_{t}^n + \textbf{l}_{t \rightarrow t+1}^n$.

\subsection{Projection matrix from camera parameters}
In computer graphics, a projection matrix is a 3x4 transformation matrix used to project three-dimensional objects onto a two-dimensional plane. It is a crucial element for football vision systems as it maps the camera's coordinate system into the real-world coordinate system, which, e.g., allows the estimate of the real-world position of players. We calculate projection matrix $P^n_t$ of $n$-th camera at time $t$  by converting a~rotation vector $\textbf{r}^n_t$ into 3x3 rotation matrix $R^n_t$ that has 3 degrees of freedom using a~Rodrigues method~\cite{gallego2015compact}. Given an intrinsic matrix $K^n$, which we assume is constant throughout the time, the 3x4 projection matrix $P^n_t$ is calculated as: 
\begin{equation}
P = K^n (R_t^n | \textbf{l}_t^n)
\label{eq:P}
\end{equation}

\subsection{3D playfield model}
\begin{figure}[ht]
\centerline{\includegraphics[width=0.48\textwidth]{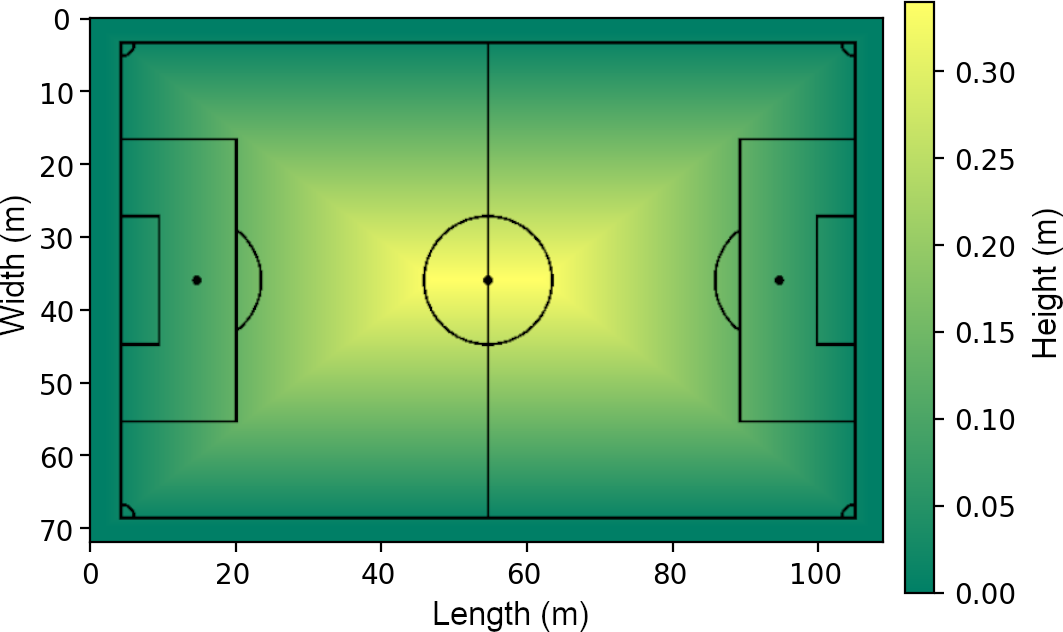}}
\caption{Our playfield model consists of four planes. The model approximates real playfields well, allowing us to achieve greater precision than baseline methods, which assume the playfield is flat.
}
\label{fig:heatmap}
\end{figure}

Based on design plans from various football fields, we model the playfield using four intersecting planes so that the middle of the playfield is elevated, as presented in Fig.~\ref{fig:heatmap}. Each plane intersects two playfield corners, which are assumed to be of height $z=0m$. Additionally, two planes intersect points $p$ and $p'$, representing edge points of the elevation. The third plane intersects $p$, and the last plane intersects $p'$. $p$, $p'$, and playfield size are hyperparameters of the playfield model.

\subsection{Evolutionary Stitched Calibration}

\begin{figure*}[ht]
\centering
\centerline{\includegraphics[width=0.95\textwidth]{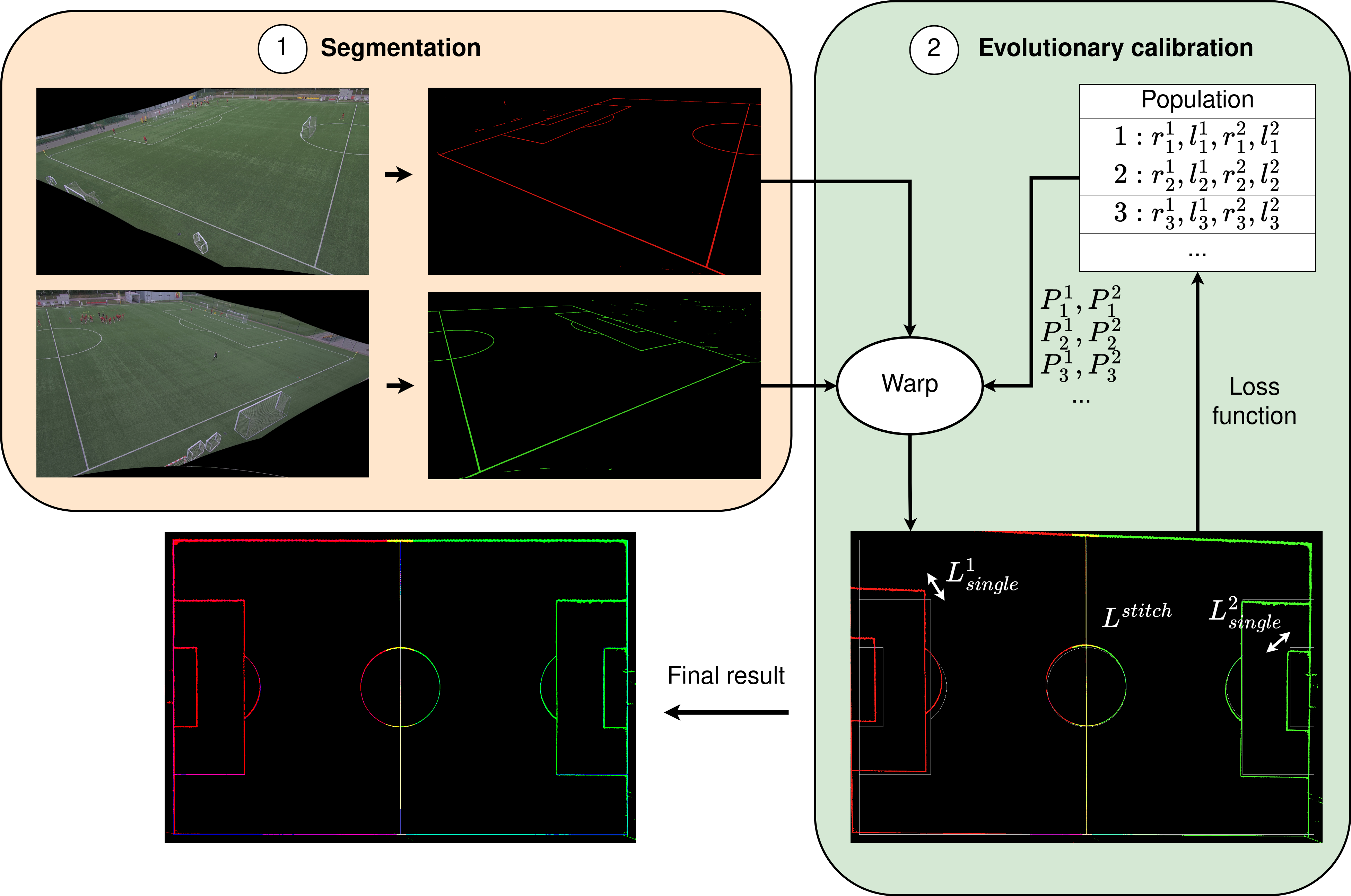}}
\caption{Our method of calibrating cameras at time $t$ consists of two phases. In the first one, we segment camera images using a deep convolutional network to find playfield lines. Next, we utilize an elitist $\mu + \lambda$ evolution strategy to find each camera's rotation and translation vectors that minimize the loss function. For $N$ cameras (here $N=2$), the loss function consists of $N+1$ summands: one per camera measuring how well the warped segmented image aligns with the playfield template and a stitch one measuring how well warped images of all cameras are aligned together. The template is represented by white field lines on a black background. To warp images, we utilize projection matrices calculated from individuals.
}
\label{fig:approach}
\end{figure*}

Our method extends ideas presented in \cite{rypesc2022sports} and improves it by enabling a 3D model of the sports field and allowing better camera stitching by jointly calibrating all cameras with a novel loss function. It consists of two phases, as presented in Fig.~\ref{fig:approach}.
The first phase is the semantic segmentation of playfield images. We utilize a deep neural network to detect playfield lines that are later used for calibration purposes. In the second stage, we reformulate the calibration process as optimization and utilize $\mu + \lambda$ evolutionary strategy to calibrate cameras.

In our approach, we assume a known size of the planar sports field, its template, and that there are $N$ cameras that share part of the play field view. We also assume we are given starting extrinsic camera parameters in the form of a~vector $\{\textbf{r}^n_t, \textbf{l}^n_t$\} for each camera $n$.

\textbf{Semantic Segmentation} The purpose of the semantic segmentation model, denoted by the function $\Psi$, is to segment an input camera image $I^n_t$ into an image $\Psi(I^n_t)$ of the same size containing only sports field lines. Segmented images are needed for the second phase of the calibration process, where they are used to calculate the value of the loss function by measuring alignment with a~blurred sports field template $T$. To obtain $T$, we combine several blurred play field template images. We blur each one of them with a~Gaussian function of a~different radius. Such transformation makes the fitness function smoother and easier to optimize~\cite{rypesc2022sports}.

For $\Psi$, we adapt U-Net architecture~\cite{ronneberger2015u}. We add a sigmoid activation function after the final layer and additional padding of 1 in encoder convolutional layers to ensure that $\Psi(I^n_t)$ is of the same size as $I^n_t$. Maintaining the image size is crucial for calibration quality. We also replace ReLU activations with LeakyReLU to decrease the number of dead neurons.

Our algorithm is designed to work with camera images segmented into playfield lines. However, other key points of the playfield can also be used. The problem of semantic segmentation of playfield lines is not trivial, as there is a strong imbalance between the background and foreground (field lines) pixels. Additionally, lines that are placed far from the camera appear smaller in the image and contribute less to cross-entropy loss functions than lines closer to the camera. To compensate for this, we train $\Psi$ utilizing focal loss~\cite{lin2017focal}.

\textbf{Evolutionary optimization} The evolution strategy we propose is based on $\mu + \lambda$ elitist evolution strategy, where in each generation $g \in {1, 2, ..., G}$ all $\mu$ parents are recombined to create an offspring of size $\lambda$. The optimization lasts $G$ generations. Individual is a vector of length $6N$ consisting of rotation and translation vectors for each camera. The starting population with $\mu$ individuals is initialized using extrinsic camera parameters from time $t$: $\textbf{r}^n_t$, $\textbf{t}^n_t$ and randomly mutating them by corrupting with a~Gaussian noise. We add  $\mathcal{N}(0, s_r)$ to rotation and $\mathcal{N}(0, s_t)$ to translation vectors respectively. $s_r$ and $s_t$ denote the strength of the rotation and translation mutation, respectively. In each generation, we create an offspring of $\lambda$ children. During recombination, we interpolate child from two parents: $[\textbf{a} * \textbf{p}_i + (\textbf{1}-\textbf{a}) * \textbf{p}_j]$ where $\textbf{a} \in \mathcal{U}(0, 1)^{6N}$, $\textbf{p}_i$ and $\textbf{p}_j$ are randomly sampled parent vectors from $P_g$. The offspring is further mutated by adding random vector $\mathcal{N}(0, s_{decay} s_r)^{3N}  \times \mathcal{N}(0, s_{decay} s_t)^{3N}$ where $s_{decay}$ denotes a~decay factor of mutation strength. Thanks to it, the strategy has improved exploration in early generations and better exploitation of found minima in the latter generations. After the mutation step is done, we calculate projection matrices for each camera for each individual following Eq.~\ref{eq:P}. Based on them, we warp segmented images into real-world coordinates to calculate the loss function $L$ for each individual. $L$ includes stitch loss $L_{stitch}$ calculated for the area of playfield visible by every camera as:
\begin{equation}
L_{stitch} = \frac{|\cap_{i=1}^N P^i(\Psi(I^i))|}{|\cup_{i=1}^N P^i(\Psi(I^i))|},
\end{equation}
where $P^i$ denotes warping camera image using the projection matrix, $\cap$ and $\cup$ represent union and intersection of images respectively, $|;|$ denotes the number of pixels in the image greater than 0.5 (segmented images are normalized to range $[0; 1]$.)  Additionally, $L$ consists of $L^i_{single}$ loss calculated for each camera $i$ as:
\begin{equation}
\label{eq:single}
L^i_{single} =  \frac{|P^i(\Psi(I^i)) \cap T|}{|P^i(\Psi(I^i)) \cup T|}
\end{equation}
Finally, $L$ is calculated  using the following formula:

\begin{equation}
    L = \lambda L_{stitch} + \frac{1 - \lambda}{N}\sum^N_{i=1} L^i_{single},
\end{equation}
where $\lambda \in [0;1]$ denotes a trade-off between the quality of the stitch and numerical accuracy of projections.

We select $\mu$ individuals with the lowest loss from the parents and offspring pool to advance to the next generation. After the last generation, the new extrinsic parameters of each camera are set to the parameters represented by the best individual.

\section{Experiments}
\subsection{Dataset}
We compare our method to existing state-of-the-art calibration methods. For this purpose, we utilize an internal football dataset of 35 sports events from 21 playfields. Each event was recorded by a pair of cameras looking at different playfield parts with a mutual intersection area around the middle line of the playfield, as presented in Fig.\ref{fig:problem}, therefore $N=2$. We randomly sample 30 images from each recording, thus giving us 2100 images of high resolution (3840x2160 pixels). We undistort them using Zhang's method~\cite{zhang2000flexible} and split this dataset into train and test sets in the proportion of 29:6.

Additionally, for each playfield, we are given its 3D model and ground truth camera parameters (intrinsic and extrinsic) obtained using Zhang\cite{zhang2000flexible} method and manual pose refinement to find a perfect stitch of camera images. We also are given a starting camera pose, denoted as Start, which is outdated for each camera, calculated with images at least one week before the event. All methods aim to find ground truth extrinsic camera parameters of each camera when initializing the algorithm with outdated matrices denoted as \emph{Start}.

\subsection{Baselines}
We compare well-established camera calibration methods to our method. Firstly, we utilize a Zhang method\cite{zhang2000flexible} where we manually find the position of key points in the images (playfield corners) and solve linear equations to obtain rotation and translation vectors. For this, we use an OpenCV~\cite{bradski2000opencv} implementation. Secondly, we utilize an Iterative Homography Network (IHN) approach\cite{cao2022iterative}, where we follow the original implementation and utilize a Siamese convolutional neural network as the feature extractor. Then, we test Optimizing Through Learned Errors (OTLE)\cite{jiang2020optimizing}, where a deep neural network (ResNet18) is used to estimate calibration error, and SGD aims to find a homography matrix that minimizes error. Finally, we utilize Sports Camera Pose Refinement (SCPR)\cite{rypesc2022sports} that assumes that playfields are flat and utilizes an evolution strategy to find camera pose that minimizes $L^i_{single}$ from Eq.\ref{eq:single} for each camera $i$.

\subsection{Metrics}
We repeat each experiment three times and report the mean and standard deviation of the following metrics:
\begin{itemize}
    \item Stitch - Measures the quality of stitch of images. It is a distance calculated in pixels on playfield images warped to bird's eye view. It is calculated between the playfield center registered and warped by the first and second camera. We set image resolution to $1\text{px} = 10\text{cm}$. 
    \item Translation error (TrE) - L2 distance in real-world (centimeters) between estimated and ground truth translation vector. Here, it is a mean calculated for a pair of cameras.
    \item Rotation error (RoE) - A mean angle (in degrees) between estimated and ground truth rotation vector.
    \item Intersection over Union (IoU) - Intersection over Union calculated between the warped playfield and the template as used in previous works~\cite{jiang2020optimizing, chen2019sports, sharma2018automated}. A value of 100\% means perfect projection between the image and real-world coordinates.
\end{itemize}

\subsection{Implementation details}
We utilize original implementations for all baseline methods except for Zhang's method, for which we utilize a well-established OpenCV~\cite{bradski2000opencv} implementation. We also set default hyperparameters for these methods.

We implement our method (ESC) in Python using PyTorch~\cite{paszke2017automatic} machine learning library. As the segmentation neural network, we utilize U-Net~\cite{ronneberger2015u} with four times fewer filters in each convolutional layer. To train it, we utilize AdamW optimizer with weight decay equal to $0.1^4$, starting learning rate of $0.1^3$ with exponential scheduling and two epochs of a warm-up. We augment data using random flips, translations, rotation, shears, scaling, and crops. For ours evolution strategy we set number of generations $G$ to $100$, $\mu=64, \lambda=128$, mutation strengts $s_t=0.1$ and $s_r=0.005$. We also set mutation decay $s_{decay}$ to 0.95.

\subsection{Results}
We provide calibration results in Tab.~\ref{table:main}. Our method is superior to baselines in terms of every metric. It achieves a stitch quality of 0.3 px, which is better than the second-best method (DHN) by 1.9 px. This shows that ESC enables superior video stitching quality, which is necessary for broadcasting sports events. Manual methods such as Zhang's (10.2 px), achieve worse stitching quality. That is because they do not take stitching into account and do not compensate for camera errors, such as imperfect image distortion being based solely on point reference. On the other hand, automatic methods such as DHN, OTLE, IHN, and SCPR do not factor in the fact that the playfield is not flat, which can explain their worse results.

ESC also outperforms baseline methods with regard to the quality of projections between the camera planes and the real-world model. It achieves 0.2 translation, 0.1 rotation, and 99.4\% IoU errors, which is better than the second-best method (SCPR) by 0.3, 0.2, and 0.3, respectively. This improvement seems to stem from removing the assumption that outdoor football playfields are flat. Therefore, for good performance of computer vision systems in sports, the model of the playfield should always be considered. 

\begin{table}[htbp]
\caption{Calibration results}
\begin{center}
\begin{tabular}{ c|c|c|c|c } 
\hline
 & \textbf{Stitch (px)} $\downarrow$ & \textbf{TrE (cm)} $\downarrow$ & \textbf{RoE ($^{\circ}$)} $\downarrow$ & \textbf{IoU (\%)} $\uparrow$\\
\hline
Start & $14.3\pm3.5$ & $16.2\pm1.3$ & $2.1\pm0.4$ & $96.2\pm0.9$ \\ 
\hline
Zhang~\cite{zhang2000flexible} & $10.2\pm1.1$ & $5.0\pm0.5$ & $1.0\pm0.1$ & $98.3\pm0.5$ \\
DHN\cite{detone2016deep} & $2.2\pm0.5$ & $12.6\pm0.6$ & $1.7\pm0.2$ & $97.1\pm0.7$ \\
OTLE\cite{jiang2020optimizing} & $8.8\pm0.9$ & $6.8\pm0.5$ & $1.3\pm0.1$ & $97.8\pm0.6$ \\
IHN~\cite{cao2022iterative} & $6.8\pm0.8$ & $2.5\pm0.4$ & $0.7\pm0.1$ & $98.7\pm0.4$ \\
SCPR\cite{rypesc2022sports} & $4.2\pm0.6$ & $0.5\pm0.1$ & $0.3\pm0.0$ & $99.1\pm0.4$ \\
ESC (ours) & $\mathbf{0.3\pm0.1}$ & $\mathbf{0.2\pm0.0}$ & $\mathbf{0.1\pm0.0}$ & $\mathbf{99.4\pm0.2}$ \\

\hline
\end{tabular}
\end{center}
\label{table:main}
\end{table}

\begin{figure}[ht]
\centerline{\includegraphics[width=0.45\textwidth]{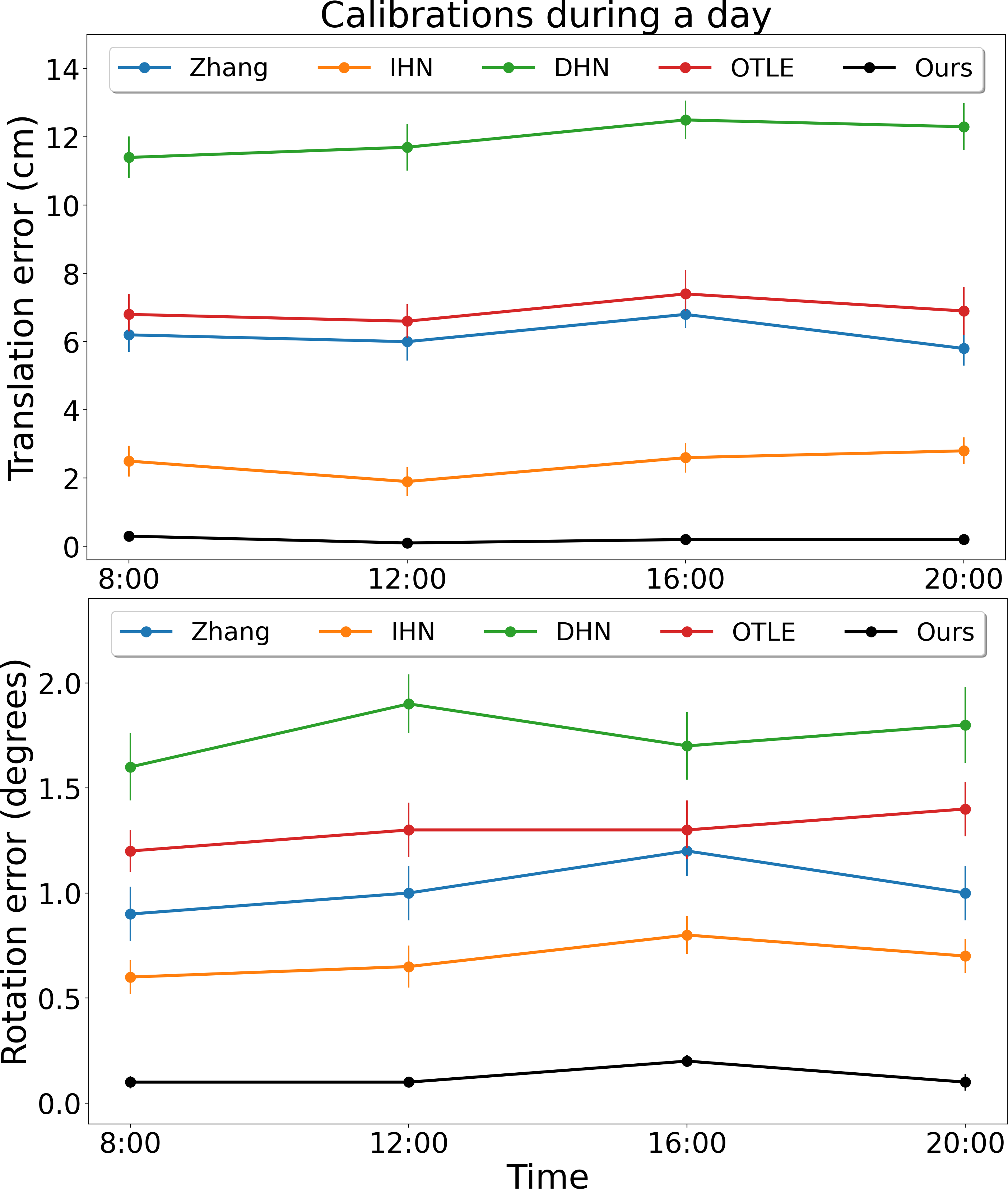}}
\caption{Evaluation of methods at different hours during the day. ESC achieves much lower translation and rotation errors compared to other approaches at any time of the day. This results in better projections and stitch quality.
}
\label{fig:during-day}
\end{figure}

Additionally, we verified ESC and baseline methods in real-life scenarios. For this purpose, we gathered data from three different sports fields at four different times of the day: 8:00, 12:00, 16:00, and 20:00. At each time, we calibrated cameras and measured translation and rotation errors. We present results in Fig.\ref{fig:during-day}. ESC achieves the lowest errors at any time of the day, proving it can deal with settings, where cameras need to be calibrated periodically during the day. IHN achieves second-best results, while DHN performs the worst. 

\begin{figure*}[ht]
\centering
\centerline{\includegraphics[width=0.95\textwidth]{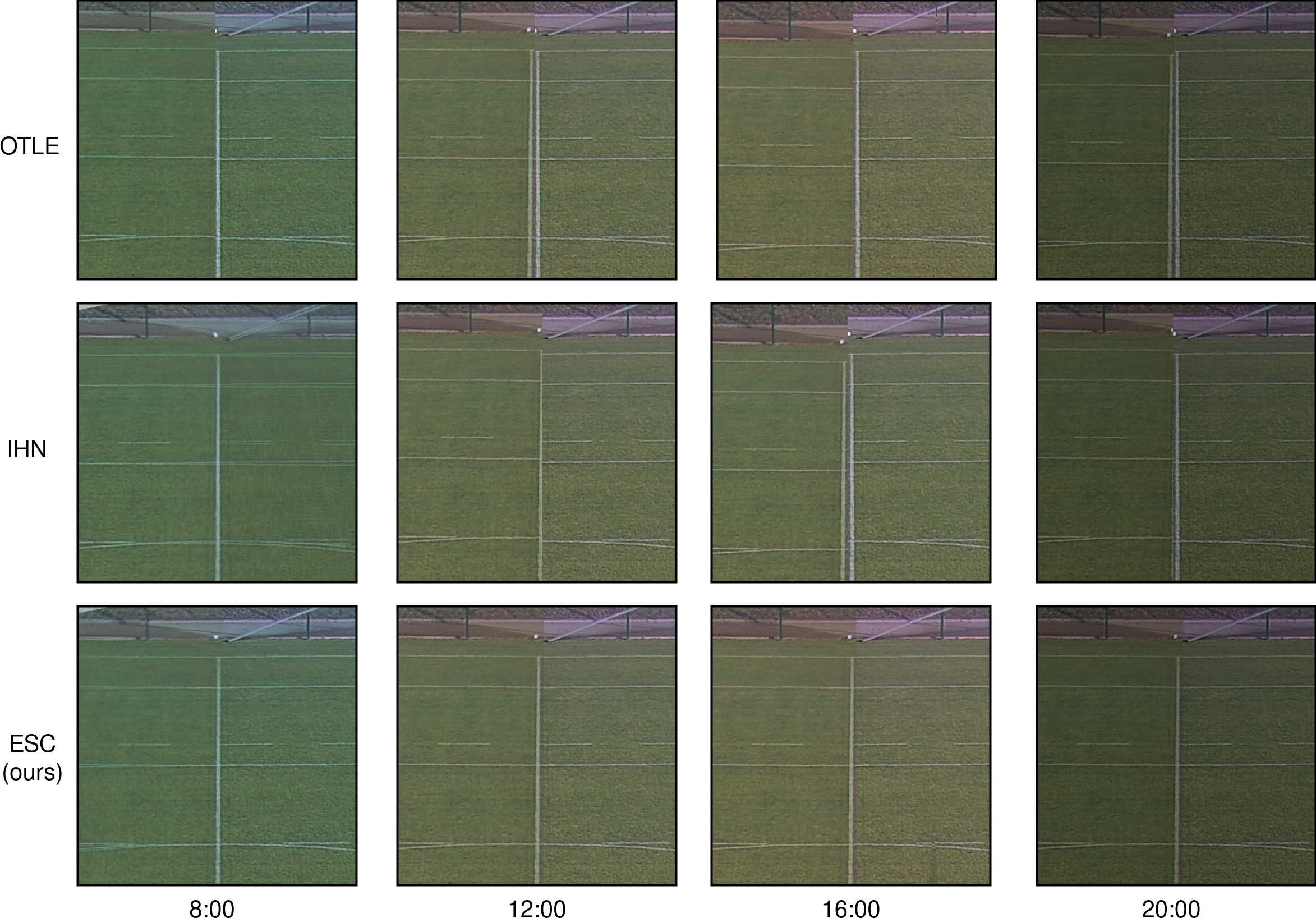}}
\caption{Stitch quality during different times of the day for different methods. ESC achieves the best results. Best viewed when zoomed in.
}
\label{fig:stitch}
\end{figure*}

As numerical superiority is not always the top priority in sports vision systems, we provide qualitative results in Fig.~\ref{fig:stitch}. We perform calibrations for different times of the day using OTLE, DHN, and ESC and look at the stitching result along the middle line based on calculated camera calibration. It is clearly visible that OTLE and DHN cannot stitch the images well; thus, they have a high stitch error. Depending on the features of the playfield and the calibration error different stitching artifacts occur. The view with imperfect calibration often results in a double middle line (e.g., OTLE at 12:00 o'clock), which is unacceptable for viewers of a football match. On the other hand, our method is able to find camera parameters that allow for a perfect stitch, proving it is a better choice for the problem.

\subsection{Analysis of ESC}

\begin{figure}[ht]
\centerline{\includegraphics[width=0.49\textwidth]{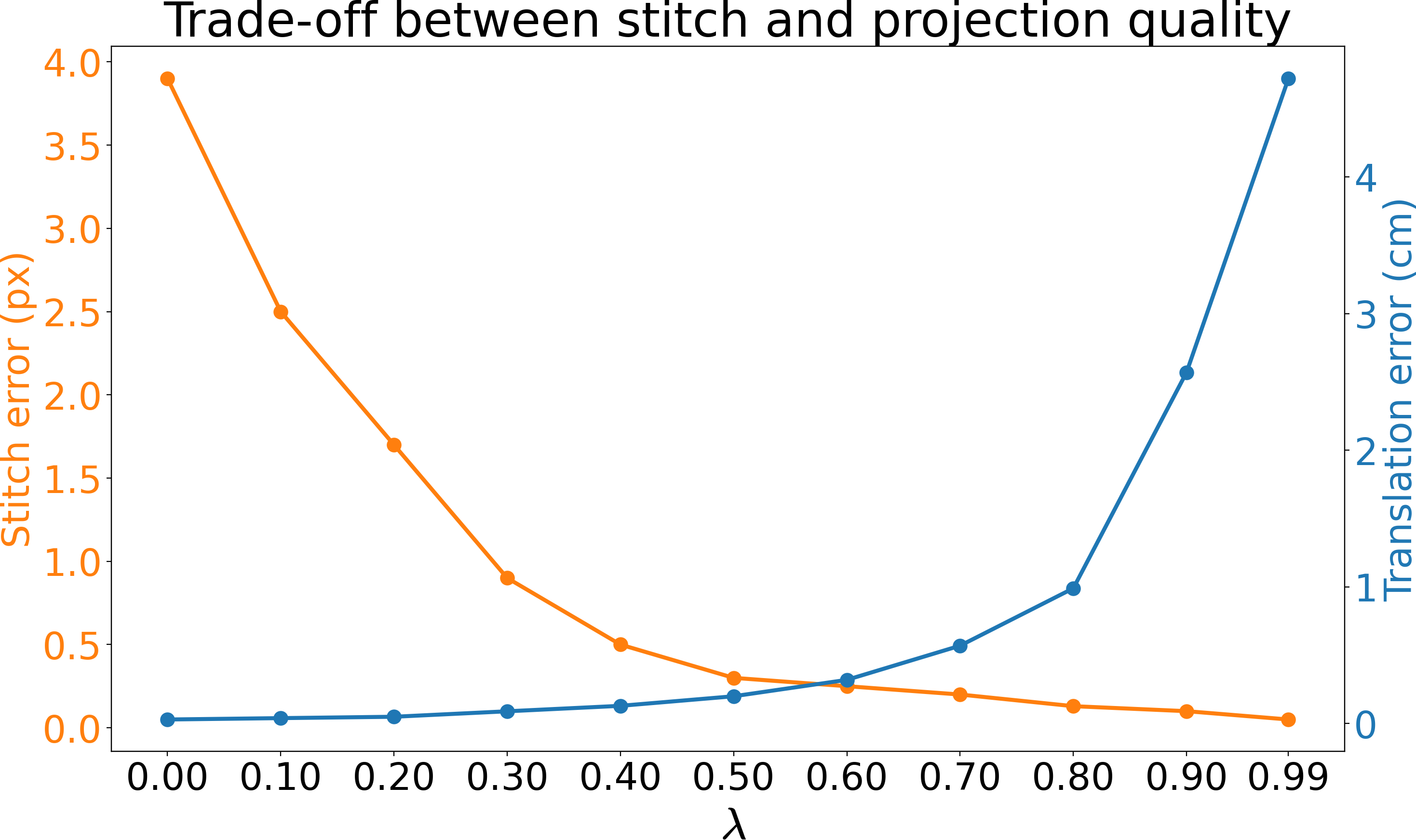}}
\caption{Depending on the task, the quality of stitch and projections from image plane to 3D world model can be adjusted using $\lambda$ parameter.
}
\label{fig:tradeoff}
\end{figure}

First, we verify how the $\lambda$ parameter affects the results achieved by our method. We report stitch quality and translation errors in Fig.~\ref{fig:tradeoff} showing their tradeoff. Here, translation error represents projection quality, as a more precise translation vector enables a more accurate projection matrix. We can see that higher $\lambda$ values increase translation error, thus decreasing the quality of projections. However, at the same time, they decrease stitch errors and overcome parallax effects and intrinsic calibration imperfections, allowing better image stitching. It is best to set $\lambda$ to values close to 0.5 for our method to achieve good-quality projections and image stitching.

Secondly, we perform an ablation study of our evolution strategy. We compare it to a version that assumes the playfield is flat and to a version that does not utilize stitch loss. We present results for these experiments in Tab.~\ref{table:ablation}. Utilizing a 3D model improves stitch quality by 0.4px and decreases stitch and rotation errors by 0.4 and 0.3, respectively. Moreover, one must include stitch loss to achieve great stitch quality, as it decreases stitch error by 3.4px. That explains the design choice of the method.

\begin{table}[ht]
\caption{Ablation study}
\begin{center}
\begin{tabular}{ c|c|c|c|c } 
\hline
 & \textbf{Stitch (px)} $\downarrow$ & \textbf{TrE (cm)} $\downarrow$ & \textbf{RoE ($^{\circ}$)} $\downarrow$ \\
\hline

ESC & $0.3\pm0.1$ & $0.2\pm0.0$ & $0.1\pm0.0$ \\
\hline
w/o 3D model & $0.7\pm0.2$ & $0.6\pm0.1$ & $0.4\pm0.1$ \\
w/o $L_{stitch}$ & $3.7\pm0.6$ & $0.2\pm0.0$ & $0.1\pm0.0$  \\

\hline
\end{tabular}
\end{center}
\label{table:ablation}
\end{table}

Finally, we verify whether our U-Net convolutional neural architecture changes improve the results. For this purpose, we test it with and without LeakyReLU activation functions and the Focal Loss cost function. To evaluate segmentation models in terms of binary classification of pixels (playfield lines detection), we utilize the popular F1 score and mean intersection over union (mIoU) metric. We present results in Tab.~\ref{table:segmentation}. Replacing the ReLU activation function with LeakyReLU improves mIoU and F1 scores by 2.3\% and 2.2\% points, respectively. Training the model with Focal Loss instead of cross entropy boosts the metrics by 5.7\% and 6.1\%, respectively. Combining these two methods improves overall results by 7.7\% and 7.8\%, which proves the design choice of our segmentation network.

\begin{table}[htbp]
\caption{Segmentation results}
\begin{center}
\begin{tabular}{ c|c|c } 
\hline
 & \textbf{mIoU (\%)} $\uparrow$ & \textbf{F1 (\%)} $\uparrow$ \\
\hline
U-Net   & $49.6\pm1.2$ & $65.8\pm1.7$  \\
with LeakyReLU  & $51.9\pm1.1$ & $68.0\pm1.9$  \\
with Focal Loss  & $55.3\pm1.3$ & $71.9\pm1.8$  \\
with both (ours)  & $\mathbf{57.3\pm1.2}$ & $\mathbf{73.6\pm1.8}$  \\
\hline

\end{tabular}
\end{center}
\label{table:segmentation}
\end{table}

\section{Conclusion}
In this work, we propose a novel method, dubbed ESC, to tackle the problem of continuous stitched camera calibration. It consists of two steps. Firstly, it segments input images using the deep convolutional network to detect playfield lines. Then, it performs evolutionary optimization to find the translation and rotation vectors of each camera that guarantee high-quality image stitch and projections. Unlike other methods, we do not assume that the playfield is flat and represent it using a 3D playfield model. That enables our method to obtain superior visual fidelity of image stitch and numerical accuracy. A series of experiments presented in the paper proved that it achieves better results than SOTA baselines: higher quality image stitching and more accurate projections. 

\noindent\textbf{Acknowledgments}
This study was prepared within realization of the Project co-funded by polish National Center of Research and Development, Ścieżka dla Mazowsza/2019.

{\small
\bibliographystyle{IEEEtran}
\bibliography{bib.bib}
}

\end{document}